\pdfoutput=1

\documentclass[11pt]{article}

\usepackage[]{ACL2023}

\usepackage{times}
\usepackage{latexsym}
\usepackage{xspace}
\usepackage{algorithm}
\usepackage[noend]{algpseudocode}
\usepackage{algpseudocode}
\algrenewcommand\algorithmicrequire{\textbf{Input:}}
\algrenewcommand\algorithmicensure{\textbf{Output:}}
\usepackage{mathtools}
\usepackage{siunitx}
\usepackage{enumitem}
\usepackage{amsfonts,amsmath,amssymb,bm,xifthen}
\usepackage{caption}
\usepackage{subcaption}
\usepackage{multirow}

\usepackage[T1]{fontenc}

\usepackage[utf8]{inputenc}

\usepackage{microtype}

\usepackage{inconsolata}
\usepackage{graphicx}

\title{Uncertainty-Aware Bootstrap Learning for Joint Extraction on Distantly-Supervised Data}

\author{
Yufei Li$^{1}$,
\xspace    
Xiao Yu$^{2}$,
\xspace    
Yanchi Liu$^{3}$,
\xspace    
Haifeng Chen$^{3}$,
\xspace    
Cong Liu$^{1}$\\
$^{1}$University of California, Riverside\xspace\xspace
$^{2}$Stellar Cyber\xspace\xspace
$^{3}$NEC Labs America\\
$^{1}$\texttt{\{yli927,congl\}@ucr.edu}, $^{2}$\texttt{xyu@stellarcyber.ai},
\\ 
$^{3}$\texttt{\{yanchi,haifeng\}@nec-labs.com}
}

\begin{document}
\maketitle

\begin{abstract}

Jointly extracting entity pairs and their relations is challenging when working on distantly-supervised data with ambiguous or noisy labels.
To mitigate such impact, we propose \textit{uncertainty-aware bootstrap learning}, which is motivated by the intuition that the higher uncertainty of an instance, the more likely the model confidence is inconsistent with the ground truths.
Specifically, we first explore instance-level data uncertainty to create an initial high-confident examples. 
Such subset serves as filtering noisy instances and facilitating the model to converge fast at the early stage.
During bootstrap learning, we propose self-ensembling as a regularizer to alleviate inter-model uncertainty produced by noisy labels. 
We further define probability variance of joint tagging probabilities to estimate inner-model parametric uncertainty, which is used to select and build up new reliable training instances for the next iteration.
Experimental results on two large datasets reveal that our approach outperforms existing strong baselines and related methods.
\end{abstract}

\section{Introduction}

Joint extraction involves extracting multiple types of entities and relations between them using a single model, which is necessary in automatic knowledge base construction~\cite{yu2019joint}. 
One way to cheaply acquire a large amount of labeled data for training joint extraction models is through distant supervision (DS)~\cite{mintz2009distant}. 
DS involves aligning a knowledge base (KB) with an unlabeled corpus using hand-crafted rules or logic constraints. 
Due to the lack of human annotators, DS brings a large proportion of noisy labels, e.g., over 30\% noisy instances in some cases~\cite{mintz2009distant}, making it impossible to learn useful features.
The noise can be either false relations due to the aforementioned rule-based matching assumption or wrong entity tags due to limited coverage over entities in open-domain KBs.

Existing distantly-supervised approaches model noise relying either on heuristics such as reinforcement learning (RL)~\cite{nooralahzadeh2019reinforcement, Hu-RL-relation-21} and adversarial learning~\cite{Chen-relation-adversarial-21}, or pattern-based methods~\cite{jia2019arnor, Shang-pattern-relation-22} to select trustable instances.
Nevertheless, these methods require designing heuristics or hand-crafted patterns which may encourage a model to leverage spurious features without considering the confidence or uncertainty of its predictions.

In response to these problems, we propose \textbf{UnBED}---\textbf{Un}certainty-aware \textbf{B}ootstrap learning for joint \textbf{E}xtraction on \textbf{D}istantly-supervised data. 
UnBED assumes that 1)~low data uncertainty indicates reliable instances using a pre-trained language model~(PLM) in the initial stage, 2) model should be aware of trustable entity and relation labels regarding its uncertainty after training.
Our bootstrap serves uncertainty as a principle to mitigate the impact of noise labels on model learning and  validate input sequences to control the number of training examples in each step.
Particularly, we quantify data uncertainty of an instance according to its \textit{winning score}~\cite{vanilla} and \textit{entropy}~\cite{Shannon-entropy}. 
We define averaged maximum probability that is estimated by a joint PLM over each token in a sequence to adapt previous techniques in joint extraction scheme.
Instances with low data uncertainty are collected to form an initial subset, which is used to tune the joint PLM tagger and facilitate fast convergence. 
Then, we define parametric uncertainty in two perspectives---inter-model and inner-model uncertainty.
The former is quantified by self-ensembling~\cite{wang-etal-2022iclr-ensemble} and serves as a regularizer to improve model robustness against noisy labels during training.
The latter is captured by probability variance in MC Dropout~\cite{Gal-16-dropout} for selecting new confident instances for the next training iteration. 
Such two-fold model uncertainties reinforce with each other to guide the model to iteratively improve its robustness and learn from reliable knowledge.

\section{Related Work}

\textbf{Joint Extraction Methods}
Joint extraction detects entities and their relations using a single model, which effectively integrates the information from both sources and therefore achieves better results in both subtasks compared to pipelined methods~\cite{zheng-etal-2017-joint}.
For example, unified methods tag entities and relation simultaneously, e.g., \cite{zheng-etal-2017-joint} proposes a novel tagging scheme which converts joint extraction to a sequence labeling problem; \cite{dai2019joint} introduces query position and sequential tagging to extract overlapping relations. 
Such methods avoid producing redundant information compared to parameter-sharing neural models~\cite{miwa2016end, gupta2016table}, and require no hand-crafted features that are used in structured systems~\cite{yu2019joint}.

To address the challenge of learning from DS, pre-trained transformers (e.g., BERT, GPT-2) have gain much attention. 
They model strong expressive context-aware representations for text sequence through multiple attention layers, and achieve state-of-the-art performance on various NLP tasks~\cite{radford2019language,devlin-etal-2019-bert,li-etal-2022-share}. 
They can be cheaply fine-tuned to solve different downstream tasks including NER and RC. 
Specifically, BERT is trained on large English corpus using masked language modeling. 
The multi-head attention weights indicate interactions between each pair of words and its hidden states integrate semantic information of the whole sentence, which are used to decode different tagging results.

\textbf{Uncertainty Methods}
Uncertainty generally comes from two sources---aleatoric uncertainty and epistemic uncertainty.
The former is also referred to as data uncertainty, describing noise inherent in the data generation.
Methods mitigating such uncertainty include data interpolation~\cite{dong-etal-2018-confidence}, winning score, and temperature scale~\cite{Guo-2017-temperature}.
The latter is also called model uncertainty, describing whether the structure choice and model parameters best describe the data distribution.
One main solution to mitigate model uncertainty is Bayesian Neural Network (BNN)~\cite{BNN} that puts a prior distribution on its weights.
To save computational cost,
Monte Carlo dropout is proposed as an approximation of variational Bayesian inference~\cite{Gal-16-dropout}, realized by training models with dropout layers and testing with stochastic inference to quantify probability variance.
Besides BNN, self-ensembling~\cite{wang-etal-2022iclr-ensemble} which measures the outputs variance between models with the same architecture has been shown effective to reduce parametric uncertainty across models.

\section{Joint Extraction Model} 

\begin{figure}
    \centering
    \includegraphics[width=0.3\textwidth]{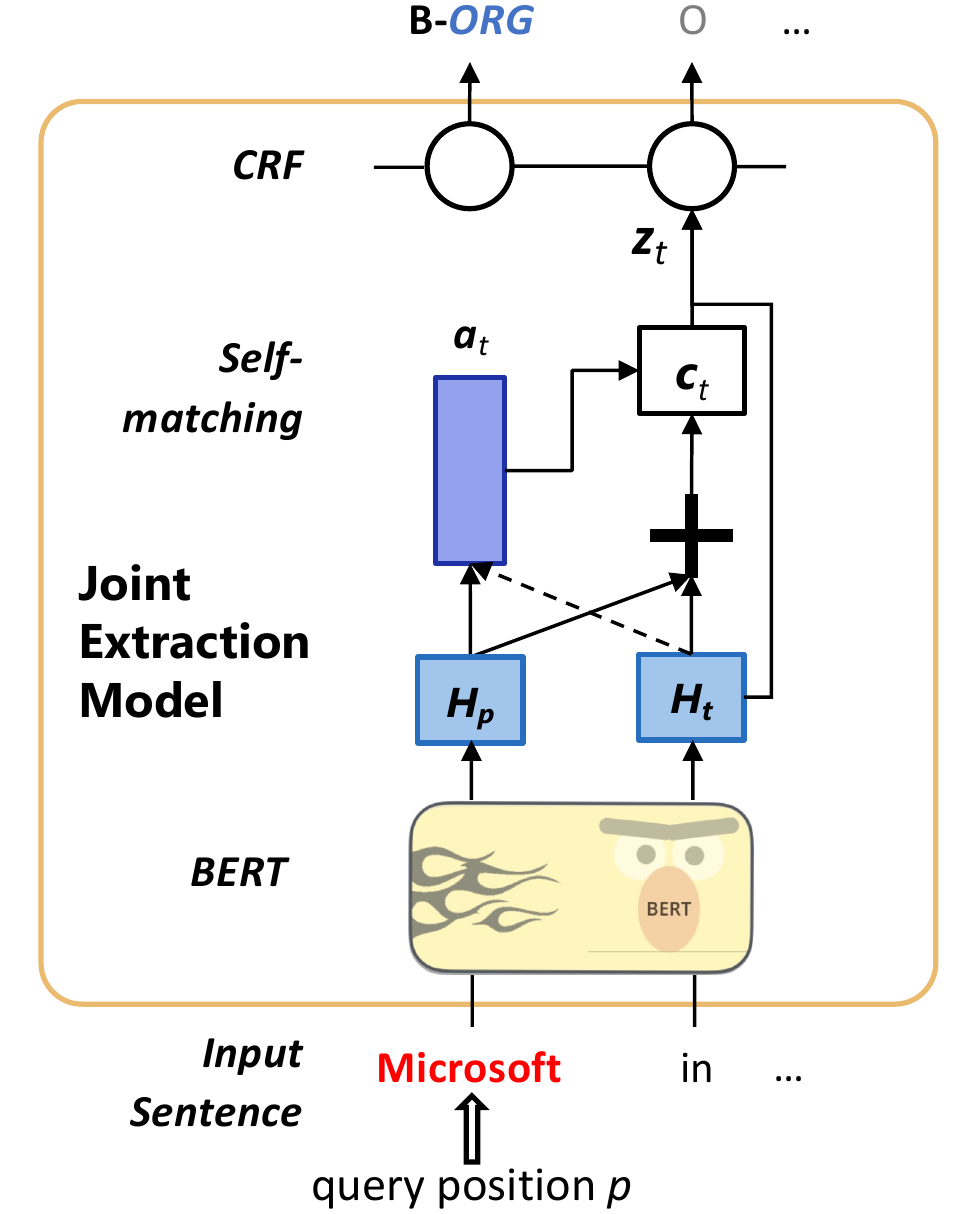}
    \caption{Joint extraction as a token classification task.}
    \label{fig:joint_model}
    \vspace{-0.3cm}
\end{figure}

\textbf{Tagging Scheme}
For an input sequence $\mathcal{X} = \left \{x_1,...,x_n \right \}$, we tag $n$ sequences according to different query position $p$ following~\cite{dai2019joint}. 
If $p$ is the start of an entity (query entity $e_1$), the sequence is an instance.
The entity type is labeled at $p$ and other entities $e_2$ which have relationship with the query entity are labeled with relation types $re$. 
The rest of tokens are labeled ``O''~(Outside), meaning they do not correspond to the query entity. 
Accordingly, we convert joint extraction into a token classification task and extract relation triplets $\left \{ e_1, re, e_2 \right \}$ in each instance, as shown in Figure~\ref{fig:joint_model}.

\textbf{Position-Attentive Encoder}
we use BERT~\cite{devlin-etal-2019-bert} to encode a sentence $\mathcal{X}$ into token-level representations $\bm{h} = \left \{ \bm{h}_1,..,\bm{h}_n \right \}$, where $\bm{h}_{i} \in \mathbb{R}^{d}$ is a $d$-dimensional vector corresponding to the $i$-th token in $\mathcal{X}$. 
For each query $p$, self-matching is applied to calculate the position-attention $\bm{a}_{t} \in \mathbb{R}^{T}$ between token at $p$ and each token at target position $t$, which compares the sentence representations against itself to collect context information~\cite{tan2018deep}. 
The produced position-aware representation $\bm{c}_{t}\in \mathbb{R}^{T\times d}$ is an attention-weighted sentence vector $\bm{c}_{t} = \bm{a}_{t}^{\top} \bm{h}$.
Finally, we concatenate $\bm{h}_t$ and $\bm{c}_{t}$ to generate position-aware and context-aware representations $\bm{u}_{t} = [\bm{h}_t| \bm{c}_{t}]$.

\textbf{CRF Decoder}~\cite{lafferty2001conditional} 
For each position-aware representation $\bm{u}_t$,  we first learn a linear transformation $\bm{z}_{t} = \bm{W}\bm{u}_{t} \in \mathbb{R}^{C}$ to represent tag scores for the $t$-th token. Here $C$ is the number of distinct tags. 
For an instance with labels $\bm{y} = \left \{y_1,...,y_n \right \}$, the decoding score $s(\bm{z}, \bm{y})$ is the sum of transition score from tag $y_t$ to tag $y_{t+1}$ plus the input score $z_t^{y_t}$. 
The conditional probability $p(\bm{y}|\bm{z})$ is the softmax over $s(\bm{z}, \bm{y})$ for all possible label sequences $\bm{y}'$. 
We maximize the log-likelihood of correct tag sequences during training $\mathcal{L}_c = \sum{\log p(\bm{y}|\bm{z})}$.

\section{Uncertainty-Aware Bootstrap Learning}
\label{sec:denoise}

\textbf{Motivation}
One of the main challenges in bootstrap learning is to evaluate the ``correctness'' of a labeled instance.
We consider this problem from an uncertainty perspective and assume instances with lower uncertainty are more likely to be correctly labeled.
In this section, we first propose instance-level data uncertainty which is used to filter noisy examples and build an initial subset.
Then, we introduce our two-fold model uncertainties which helps iteratively mitigate DS effect and build up trustable examples during bootstrap learning.

\subsection{Data Uncertainty} 

Presenting examples in an easy-to-hard order at different training stages can benefit models~\cite{Platanios-naacl-19,zhou-etal-2020-uncertainty}, we propose data uncertainty as a way to quantify the ``hardness'' of an instance.
To better estimate the data uncertainty, we use pre-trained language models~(PLMs) to generate tag probability for each token in a sequence.
Our intuition is that higher uncertain inputs are ``harder'' to be generated by a PLM, as it already has rationales of language.    
Accordingly, we propose two data uncertainties, which can be used individually or combined together:

\textbf{Winning Score (WS)} The maximum softmax probability reflects data uncertainty of an input \cite{vanilla}.
Given an input instance $\mathcal{I}=\left \{x_1,...,x_n \right \}$, we define data uncertainty $u^{d}(\mathcal{I})$ as the minus averaged token classification winning score:

\begin{equation}
\resizebox{0.85\hsize}{!}{%
    $u^{d}(\mathcal{I}) = -\frac{1}{n}\sum_{t=1}^{n}{\max_{c\in [1, C]}P(y_t = c|x_t)}$
}
\end{equation}

\textbf{Entropy} Shannon entropy~\cite{Shannon-entropy} is widely used to reflect information uncertainty.
We propose data uncertainty $u^{d}(\mathcal{I})$ as the averaged token classification entropy:

\begin{equation}
\resizebox{0.99\hsize}{!}{%
    $u^{d}(\mathcal{I}) = \frac{1}{n}\sum_{t=1}^{n}{\sum_{c=1}^{C}P(y_t=c|x_t)\log P(y_t=c|x_t)}$
}
\end{equation}

We filter out examples with high uncertainty scores and build an initial subset with ``simple'' examples.
At the early training stage, a model is not aware of what a decent distribution $P(y|x)$ should be, thus data uncertainty facilitates it to converge fast by tuning on a fairly ``simple'' subset.

\subsection{Model Uncertainty}

In our bootstrap learning, we define model uncertainty, i.e., epistemic uncertainty~\cite{Kendall-nips-17}, to measure whether model parameters can best describe the data distribution following~\cite{zhou-etal-2020-uncertainty}.
A small model uncertainty indicates the model is confident that the current training data has been well learned~\cite{wang-etal-2019}.
We adopt Monte Carlo Dropout~\cite{Gal-16-dropout} to approximate Bayesian inference which captures inner-model parametric uncertainty.
Specifically, we perform $K$ forward passes through our joint model.
In each pass, part of network neurons $\theta$ are randomly deactivated.
Finally, we yield $K$ samples on model parameters \{$\hat{\theta}_1, ..., \hat{\theta}_K$\}.
We use the averaged token classification \textbf{Probability Variance (PV)}~\cite{shelmanov-etal-2021-certain} over all tags for instance $\mathcal{I}$:

\begin{equation}
\resizebox{0.99\hsize}{!}{%
    $u^{m}(\theta) =
    \frac{1}{n}\sum_{t=1}^{n}\sum_{c=1}^{C}\text{Var}\left [P(y_t=c|x_t, \hat{\theta}_k) \right]_{k=1}^K$
}
\end{equation}

where $\text{Var}[.]$ is the variance of distribution over the $K$ passes following the common settings in ~\cite{dong-etal-2018-confidence, Xiao-aaai-19}.
Accordingly, model is aware of its confidence over each instance and how likely the label is noisy.

\subsection{Training Strategy}

\begin{algorithm}[t]
\caption{Bootstrap Learning}
\label{alg:bootstrap}
\begin{algorithmic}[1]
\Require Original dataset $\mathcal{D}=\left \{ (\mathcal{I}^n, y^n) \right \}_{n=1}^N$, two joint models $f_{1}$, $f_{2}$ with parameters $\theta_1$, $\theta_2$;

\State Compute data uncertainty $u^d(\mathcal{I})$ for each instance $\mathcal{I}$ in $\mathcal{D}$;
\State Initial dataset $\mathcal{C}$ $\leftarrow$ Select data pairs $(\mathcal{I}^n, y^n)$ such that $u^d(\mathcal{I}) < \tau^d$ from $\mathcal{D}$;
\For{\textit{epoch} $e=1,...$}
    \State Train $f_{1}$, $f_{2}$ on $\mathcal{C}$ using Eq.~(\ref{eq: loss});
    \State Calculate model uncertainty $u^m(\theta_1)$ on $\mathcal{D}$;
    \State $\mathcal{C}$ $\leftarrow$ Select data pairs $(\mathcal{I}^n, y^n)$ such that $u^m(\mathcal{I};\theta_1) < \tau^m$ from $\mathcal{D}$;
    \EndFor
\end{algorithmic}
\end{algorithm}

\textbf{Uncertainty-Aware Loss}
Besides MC Dropout which measures parametric uncertainty within a model, we also consider mitigating parametric uncertainty between models to stabilize the weights during training. Specifically, we use self-ensembling~\cite{he-etal-2020-towards, wang-etal-2022iclr-ensemble} to calculate the loss between the same models to improve model robustness and reduce the label noise effect on model performance. 

We create another joint model with identical framework, e.g., architecture, loss functions, hyperparameters, and compute a self-ensemble loss $\mathcal{L}_e$ to minimize the difference between two outputs from the two models regarding the same inputs:

\begin{equation}
    \mathcal{L}_e = \sum KL(f(\mathcal{I};\theta_1), f(\mathcal{I};\theta_2))
\end{equation}

where $KL(.)$ is the Kullback-Leibler divergence between two probabilistic distributions, $\theta_1$, $\theta_2$ denote the parameters of first and second models. 
We formulate our final uncertainty-aware objective $\mathcal{L}$ as the sum of CRF and self-ensemble loss:

\begin{equation}
\label{eq: loss}
    \mathcal{L}=\mathcal{L}_c + \alpha \mathcal{L}_e
\end{equation}

where $\alpha$ denotes the weight of self-ensembling, and $\mathcal{L}_c$ means the token classification loss.

\begin{table*}[]
    \centering
    \begin{tabular}{l|ccc|ccc}
        \hline
        \multirow{2}{*}{\textbf{Method}} & \multicolumn{3}{c|}{\textbf{NYT}} & \multicolumn{3}{c}{\textbf{Wiki-KBP}}\\
        \cline{2-7}
        
        &\textbf{Prec.} & \textbf{Rec.} & \textbf{F1}&\textbf{Prec.} & \textbf{Rec.} & \textbf{F1}  \\
        \hline
        ARNOR~\cite{jia2019arnor} & 0.588 & 0.614 & 0.600 & 0.402 & 0.471 & 0.434 \\
        PURE~\cite{zhong-chen-2021-frustratingly} & 0.536 & 0.664 & 0.593 & 0.395 & 0.433& 0.413 \\
        FAN~\cite{hao-etal-2021-knowing} & 0.579 & 0.646 & 0.611 & 0.391 & 0.467 & 0.426 \\
        \textbf{UnBED-WS} & \textbf{0.662} & 0.730 & \textbf{0.694} & \textbf{0.429} & 0.501 & \textbf{0.462} \\
        \textbf{UnBED-Entropy} & 0.651 & \textbf{0.741} & 0.693 & 0.422 & \textbf{0.509} & 0.461 \\
        \hline
        
    \end{tabular}
    \caption{Evaluation results on NYT and Wiki-KBP datasets. \textbf{Bold} numbers denote the best metrics. UnBED-WS and UnBED-Entropy denote UnBED with winning score and entropy as the data uncertainty, respectively.}
    \label{tab:main_results}
\end{table*}

\textbf{Bootstrap Learning Procedure}
To mitigate the DS effect on model performance, we propose a two-fold bootstrap learning strategy (see Algorithm~\ref{alg:bootstrap}).
Specifically, we first apply data uncertainty to filter ``harder'' examples and redistribute a reliable initial training data $\mathcal{M}$. 
Then, we iteratively feed examples following an easy-to-hard order to the model.
In each training iteration, we regularize the joint model with self-ensembling loss to reduce the impact of noisy labels on model parameters.
Then we use probability variance to select new confident training instances $\mathcal{D}'$ that can be explained by the model as the next training inputs. 
The more certain examples are selected, the more likely the model will learn beneficial information and will converge faster.
We repeat the above procedure until the F1 score on the validation set converges.

\section{Experiments}

\subsection{Setup}

We evaluate the performance of UnBED on two datasets, NYT and Wiki-KBP.
The NYT~\cite{riedel2010modeling} dataset collects news from New York Times and its training data is automatically labeled by DS. 
We use the revised test dataset~\cite{jia2019arnor} that is manually annotated to ensure quality. 
The Wiki-KBP~\cite{ling2012fine} dataset collects articles from Wikipedia. 
Its training data is labeled by DS~\cite{liu-etal-2017-heterogeneous}, and the test set is manually annotated~\cite{KBP_slot}.

We compare UnBED with the following baselines:
\textbf{ARNOR}~\cite{jia2019arnor}, a pattern-based method to reduce noise for distantly-supervised triplet extraction. 
\textbf{PURE}~\cite{zhong-chen-2021-frustratingly}, a pipeline approach that uses pre-trained BERT entity model to first recognize entities and then employs a relation model to detect underlying relations.
\textbf{FAN}~\cite{hao-etal-2021-knowing}, an adversarial method including a transformers encoder to reduce noise for distantly-supervised triplet extraction.

\textbf{Evaluation} 
We evaluate the extracted triplets for each sentence based on Precision (Prec.), Recall (Rec.), and F1. 
A triplet $\left \{ e_1, re, e_2\right \}$ is marked correct if the relation type $re$, two entities $e_1$, $e_2$ are all correct.  
We build a validation set by randomly sampling 10\% sentences from the test set.

\textbf{Implementation Details}
We use Hugging Face \textit{bert-large-uncased}~\cite{devlin-etal-2019-bert} pre-trained model as backbone. 
For ARNOR, the hidden vector size is set to 300. 
In regularization training, we find optimal parameters $\alpha$ as 1 for both datasets. 
We implement UnBED and all baselines in PyTorch, with Adam optimizer, initial learning rate 10$^{-5}$, dropout rate 0.1, and batch size 8. 
For initial subset configuration, we choose data uncertainty threshold 0.5. 
For bootstrap learning, an empirical model uncertainty threshold is set to 0.6 with the best validation F1.

\subsection{Overall Results}

As shown in Table~\ref{tab:main_results}, UnBED significantly outperforms all baselines in precision and F1 metric. 
Specifically, UnBED achieves 8\% F1 improvement on NYT (3\% on Wiki-KBP) over denoising approaches---ARNOR and FAN. 
Our approach also outperforms baselines using pre-trained transformers (PURE and FAN), showing that uncertainty-aware bootstrap learning effectively reduces the impact of noisy labels.

\subsection{Further Analysis}

\begin{figure}[]
    \centering
    \includegraphics[width=0.45\textwidth]{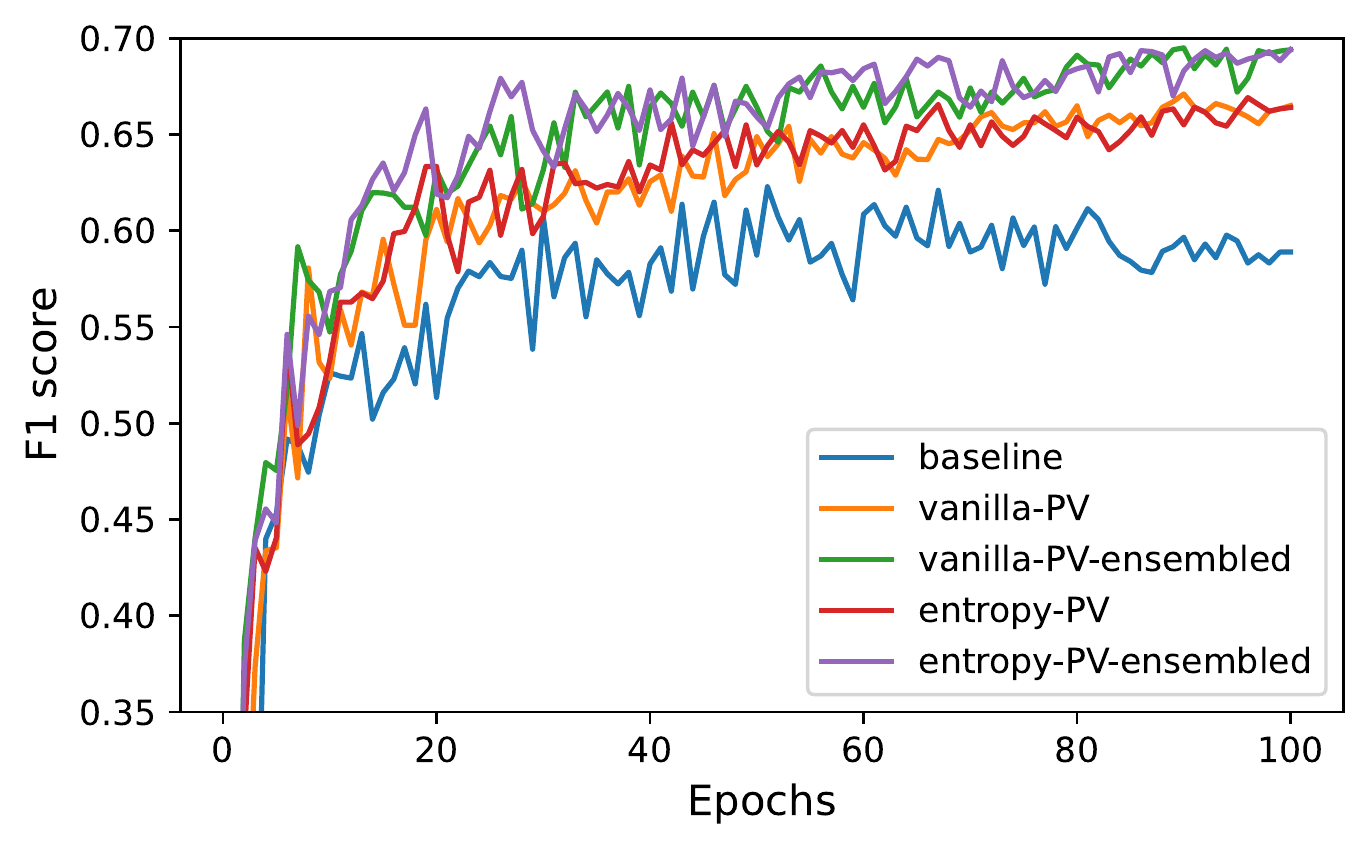}
    \caption{F1 score vs. Epochs under different settings.
    Vanilla-PV-ensembled denotes UnBED-WS, and entropy-PV-ensembled denotes UnBED-Entropy.}
    \label{fig:f1}
\end{figure}

We analyze the functionality of different components in Figure~\ref{fig:f1}.
We observe that both the entropy-PV and vanilla-PV outperform the baseline (joint model directly trained on the original DS dataset) in terms of F1 (5$\sim$7\% increase), demonstrating the effect of filtering noisy labels and selecting trustable instance using probability variance.
Besides, self-ensembling further enhances the performance in later training stage (2$\sim$4 F1 increase), proving that mitigating the inter-model uncertainty benefits model robustness against noisy labels.

\section{Conclusions}
\label{sec:conclusions}

In this paper, we propose a novel uncertainty-aware bootstrap learning framework for distantly-supervised joint extraction.
Specifically, we define data uncertainty in generally token classification to filter out highly-error-prone instances and build an initial high-confident subset, which is used to tune the joint extraction model for fast convergence.
We then propose a two-fold bootstrap learning procedure which iteratively mitigates the DS impact on model robustness and selects new trustable training instances.
Experimental results on two benchmark datasets show that UnBED significantly outperforms other denoising techniques.

\section*{Limitations}

In this work we propose an uncertainty-aware bootstrap learning framework for joint extraction. 
Though it achieves state-of-the-art performance compared to other denoising techniques, UnBED requires large training resources considering the 
ensemble loss calculated between two large PLMs and the probability variance calculated on the PLM joint extraction model.
In our future work, we hope to incorporate pruning techniques during training to improve the efficiency.
We will also consider more complex relations between entities, e.g., relations beyond the sentence boundary, to fit in real-world information extraction scenarios.


\section*{Acknowledgements}
This work was supported by NSF CNS 2135625, CPS 2038727, CNS Career 1750263, and a Darpa Shell grant.

\bibliography{anthology,custom}

\begin{thebibliography}{36}
\expandafter\ifx\csname natexlab\endcsname\relax\def\natexlab#1{#1}\fi

\bibitem[{Chen et~al.(2021)Chen, Shi, Liu, Tang, Shao, Chen, and
  Zhuang}]{Chen-relation-adversarial-21}
Tao Chen, Haochen Shi, Liyuan Liu, Siliang Tang, Jian Shao, Zhigang Chen, and
  Yueting Zhuang. 2021.
\newblock \href {https://ojs.aaai.org/index.php/AAAI/article/view/17501}
  {Empower distantly supervised relation extraction with collaborative
  adversarial training}.
\newblock In \emph{Thirty-Fifth {AAAI} Conference on Artificial Intelligence,
  {AAAI} 2021, Thirty-Third Conference on Innovative Applications of Artificial
  Intelligence, {IAAI} 2021, The Eleventh Symposium on Educational Advances in
  Artificial Intelligence, {EAAI} 2021, Virtual Event, February 2-9, 2021},
  pages 12675--12682. {AAAI} Press.

\bibitem[{Dai et~al.(2019)Dai, Xiao, Lyu, Dou, She, and Wang}]{dai2019joint}
Dai Dai, Xinyan Xiao, Yajuan Lyu, Shan Dou, Qiaoqiao She, and Haifeng Wang.
  2019.
\newblock \href {https://doi.org/10.1609/aaai.v33i01.33016300} {Joint
  extraction of entities and overlapping relations using position-attentive
  sequence labeling}.
\newblock In \emph{The Thirty-Third {AAAI} Conference on Artificial
  Intelligence, {AAAI} 2019, The Thirty-First Innovative Applications of
  Artificial Intelligence Conference, {IAAI} 2019, The Ninth {AAAI} Symposium
  on Educational Advances in Artificial Intelligence, {EAAI} 2019, Honolulu,
  Hawaii, USA, January 27 - February 1, 2019}, pages 6300--6308. {AAAI} Press.

\bibitem[{Devlin et~al.(2019)Devlin, Chang, Lee, and
  Toutanova}]{devlin-etal-2019-bert}
Jacob Devlin, Ming-Wei Chang, Kenton Lee, and Kristina Toutanova. 2019.
\newblock \href {https://doi.org/10.18653/v1/N19-1423} {{BERT}: Pre-training of
  deep bidirectional transformers for language understanding}.
\newblock In \emph{Proceedings of the 2019 Conference of the North {A}merican
  Chapter of the Association for Computational Linguistics: Human Language
  Technologies, Volume 1 (Long and Short Papers)}, pages 4171--4186,
  Minneapolis, Minnesota. Association for Computational Linguistics.

\bibitem[{Dong et~al.(2018)Dong, Quirk, and Lapata}]{dong-etal-2018-confidence}
Li~Dong, Chris Quirk, and Mirella Lapata. 2018.
\newblock \href {https://doi.org/10.18653/v1/P18-1069} {Confidence modeling for
  neural semantic parsing}.
\newblock In \emph{Proceedings of the 56th Annual Meeting of the Association
  for Computational Linguistics (Volume 1: Long Papers)}, pages 743--753,
  Melbourne, Australia. Association for Computational Linguistics.

\bibitem[{Ellis et~al.(2013)Ellis, Getman, Mott, Li, Griffitt, Strassel, and
  Wright}]{KBP_slot}
Joe Ellis, Jeremy Getman, Justin Mott, Xuansong Li, Kira Griffitt, Stephanie~M.
  Strassel, and Jonathan Wright. 2013.
\newblock \href
  {https://tac.nist.gov/publications/2013/additional.papers/KBP2013\_annotation\_overview.TAC2013.proceedings.pdf}
  {Linguistic resources for 2013 knowledge base population evaluations}.
\newblock In \emph{Proceedings of the Sixth Text Analysis Conference, {TAC}
  2013, Gaithersburg, Maryland, USA, November 18-19, 2013}. {NIST}.

\bibitem[{Gal and Ghahramani(2016)}]{Gal-16-dropout}
Yarin Gal and Zoubin Ghahramani. 2016.
\newblock \href {http://proceedings.mlr.press/v48/gal16.html} {Dropout as a
  bayesian approximation: Representing model uncertainty in deep learning}.
\newblock In \emph{Proceedings of the 33nd International Conference on Machine
  Learning, {ICML} 2016, New York City, NY, USA, June 19-24, 2016}, volume~48
  of \emph{{JMLR} Workshop and Conference Proceedings}, pages 1050--1059.
  JMLR.org.

\bibitem[{Guo et~al.(2017)Guo, Pleiss, Sun, and
  Weinberger}]{Guo-2017-temperature}
Chuan Guo, Geoff Pleiss, Yu~Sun, and Kilian~Q. Weinberger. 2017.
\newblock \href {http://proceedings.mlr.press/v70/guo17a.html} {On calibration
  of modern neural networks}.
\newblock In \emph{Proceedings of the 34th International Conference on Machine
  Learning, {ICML} 2017, Sydney, NSW, Australia, 6-11 August 2017}, volume~70
  of \emph{Proceedings of Machine Learning Research}, pages 1321--1330. {PMLR}.

\bibitem[{Gupta et~al.(2016)Gupta, Sch{\"u}tze, and Andrassy}]{gupta2016table}
Pankaj Gupta, Hinrich Sch{\"u}tze, and Bernt Andrassy. 2016.
\newblock \href {https://aclanthology.org/C16-1239} {Table filling multi-task
  recurrent neural network for joint entity and relation extraction}.
\newblock In \emph{Proceedings of {COLING} 2016, the 26th International
  Conference on Computational Linguistics: Technical Papers}, pages 2537--2547,
  Osaka, Japan. The COLING 2016 Organizing Committee.

\bibitem[{Hao et~al.(2021)Hao, Yu, and Hu}]{hao-etal-2021-knowing}
Kailong Hao, Botao Yu, and Wei Hu. 2021.
\newblock \href {https://doi.org/10.18653/v1/2021.emnlp-main.761} {Knowing
  false negatives: An adversarial training method for distantly supervised
  relation extraction}.
\newblock In \emph{Proceedings of the 2021 Conference on Empirical Methods in
  Natural Language Processing}, pages 9661--9672, Online and Punta Cana,
  Dominican Republic. Association for Computational Linguistics.

\bibitem[{He et~al.(2020)He, Zhang, Lei, Chen, Chen, Alhamadani, Xiao, and
  Lu}]{he-etal-2020-towards}
Jianfeng He, Xuchao Zhang, Shuo Lei, Zhiqian Chen, Fanglan Chen, Abdulaziz
  Alhamadani, Bei Xiao, and ChangTien Lu. 2020.
\newblock \href {https://doi.org/10.18653/v1/2020.emnlp-main.671} {Towards more
  accurate uncertainty estimation in text classification}.
\newblock In \emph{Proceedings of the 2020 Conference on Empirical Methods in
  Natural Language Processing (EMNLP)}, pages 8362--8372, Online. Association
  for Computational Linguistics.

\bibitem[{Hendrycks and Gimpel(2017)}]{vanilla}
Dan Hendrycks and Kevin Gimpel. 2017.
\newblock \href {https://openreview.net/forum?id=Hkg4TI9xl} {A baseline for
  detecting misclassified and out-of-distribution examples in neural networks}.
\newblock In \emph{5th International Conference on Learning Representations,
  {ICLR} 2017, Toulon, France, April 24-26, 2017, Conference Track
  Proceedings}. OpenReview.net.

\bibitem[{Hu et~al.(2021)Hu, Zhang, Yang, Li, Lin, Wen, and
  Yu}]{Hu-RL-relation-21}
Xuming Hu, Chenwei Zhang, Yawen Yang, Xiaohe Li, Li~Lin, Lijie Wen, and
  Philip~S. Yu. 2021.
\newblock \href {https://doi.org/10.18653/v1/2021.emnlp-main.216} {Gradient
  imitation reinforcement learning for low resource relation extraction}.
\newblock In \emph{Proceedings of the 2021 Conference on Empirical Methods in
  Natural Language Processing, {EMNLP} 2021, Virtual Event / Punta Cana,
  Dominican Republic, 7-11 November, 2021}, pages 2737--2746. Association for
  Computational Linguistics.

\bibitem[{Jia et~al.(2019)Jia, Dai, Xiao, and Wu}]{jia2019arnor}
Wei Jia, Dai Dai, Xinyan Xiao, and Hua Wu. 2019.
\newblock \href {https://doi.org/10.18653/v1/P19-1135} {{ARNOR}: Attention
  regularization based noise reduction for distant supervision relation
  classification}.
\newblock In \emph{Proceedings of the 57th Annual Meeting of the Association
  for Computational Linguistics}, pages 1399--1408, Florence, Italy.
  Association for Computational Linguistics.

\bibitem[{Kendall and Gal(2017)}]{Kendall-nips-17}
Alex Kendall and Yarin Gal. 2017.
\newblock \href
  {https://proceedings.neurips.cc/paper/2017/hash/2650d6089a6d640c5e85b2b88265dc2b-Abstract.html}
  {What uncertainties do we need in bayesian deep learning for computer
  vision?}
\newblock In \emph{Advances in Neural Information Processing Systems 30: Annual
  Conference on Neural Information Processing Systems 2017, December 4-9, 2017,
  Long Beach, CA, {USA}}, pages 5574--5584.

\bibitem[{Klein et~al.(2017)Klein, Falkner, Springenberg, and Hutter}]{BNN}
Aaron Klein, Stefan Falkner, Jost~Tobias Springenberg, and Frank Hutter. 2017.
\newblock \href {https://openreview.net/forum?id=S11KBYclx} {Learning curve
  prediction with bayesian neural networks}.
\newblock In \emph{5th International Conference on Learning Representations,
  {ICLR} 2017, Toulon, France, April 24-26, 2017, Conference Track
  Proceedings}. OpenReview.net.

\bibitem[{Lafferty et~al.(2001)Lafferty, McCallum, and
  Pereira}]{lafferty2001conditional}
John~D. Lafferty, Andrew McCallum, and Fernando C.~N. Pereira. 2001.
\newblock Conditional random fields: Probabilistic models for segmenting and
  labeling sequence data.
\newblock In \emph{Proceedings of the Eighteenth International Conference on
  Machine Learning}, ICML '01, page 282–289, San Francisco, CA, USA. Morgan
  Kaufmann Publishers Inc.

\bibitem[{Li et~al.(2022)Li, Li, Ni, and McAuley}]{li-etal-2022-share}
Shuyang Li, Yufei Li, Jianmo Ni, and Julian McAuley. 2022.
\newblock \href {https://aclanthology.org/2022.emnlp-main.761} {{SHARE}: a
  system for hierarchical assistive recipe editing}.
\newblock In \emph{Proceedings of the 2022 Conference on Empirical Methods in
  Natural Language Processing}, pages 11077--11090, Abu Dhabi, United Arab
  Emirates. Association for Computational Linguistics.

\bibitem[{Ling and Weld(2012)}]{ling2012fine}
Xiao Ling and Daniel~S. Weld. 2012.
\newblock \href {http://www.aaai.org/ocs/index.php/AAAI/AAAI12/paper/view/5152}
  {Fine-grained entity recognition}.
\newblock In \emph{Proceedings of the Twenty-Sixth {AAAI} Conference on
  Artificial Intelligence, July 22-26, 2012, Toronto, Ontario, Canada}. {AAAI}
  Press.

\bibitem[{Liu et~al.(2017)Liu, Ren, Zhu, Zhi, Gui, Ji, and
  Han}]{liu-etal-2017-heterogeneous}
Liyuan Liu, Xiang Ren, Qi~Zhu, Shi Zhi, Huan Gui, Heng Ji, and Jiawei Han.
  2017.
\newblock \href {https://doi.org/10.18653/v1/D17-1005} {Heterogeneous
  supervision for relation extraction: A representation learning approach}.
\newblock In \emph{Proceedings of the 2017 Conference on Empirical Methods in
  Natural Language Processing}, pages 46--56, Copenhagen, Denmark. Association
  for Computational Linguistics.

\bibitem[{Mintz et~al.(2009)Mintz, Bills, Snow, and
  Jurafsky}]{mintz2009distant}
Mike Mintz, Steven Bills, Rion Snow, and Daniel Jurafsky. 2009.
\newblock \href {https://aclanthology.org/P09-1113} {Distant supervision for
  relation extraction without labeled data}.
\newblock In \emph{Proceedings of the Joint Conference of the 47th Annual
  Meeting of the {ACL} and the 4th International Joint Conference on Natural
  Language Processing of the {AFNLP}}, pages 1003--1011, Suntec, Singapore.
  Association for Computational Linguistics.

\bibitem[{Miwa and Bansal(2016)}]{miwa2016end}
Makoto Miwa and Mohit Bansal. 2016.
\newblock \href {https://doi.org/10.18653/v1/p16-1105} {End-to-end relation
  extraction using lstms on sequences and tree structures}.
\newblock In \emph{Proceedings of the 54th Annual Meeting of the Association
  for Computational Linguistics, {ACL} 2016, August 7-12, 2016, Berlin,
  Germany, Volume 1: Long Papers}. The Association for Computer Linguistics.

\bibitem[{Nooralahzadeh et~al.(2019)Nooralahzadeh, L{\o}nning, and
  {\O}vrelid}]{nooralahzadeh2019reinforcement}
Farhad Nooralahzadeh, Jan~Tore L{\o}nning, and Lilja {\O}vrelid. 2019.
\newblock \href {https://doi.org/10.18653/v1/D19-6125} {Reinforcement-based
  denoising of distantly supervised {NER} with partial annotation}.
\newblock In \emph{Proceedings of the 2nd Workshop on Deep Learning Approaches
  for Low-Resource NLP, DeepLo@EMNLP-IJCNLP 2019, Hong Kong, China, November 3,
  2019}, pages 225--233. Association for Computational Linguistics.

\bibitem[{Platanios et~al.(2019)Platanios, Stretcu, Neubig, P{\'{o}}czos, and
  Mitchell}]{Platanios-naacl-19}
Emmanouil~Antonios Platanios, Otilia Stretcu, Graham Neubig, Barnab{\'{a}}s
  P{\'{o}}czos, and Tom~M. Mitchell. 2019.
\newblock \href {https://doi.org/10.18653/v1/n19-1119} {Competence-based
  curriculum learning for neural machine translation}.
\newblock In \emph{Proceedings of the 2019 Conference of the North American
  Chapter of the Association for Computational Linguistics: Human Language
  Technologies, {NAACL-HLT} 2019, Minneapolis, MN, USA, June 2-7, 2019, Volume
  1 (Long and Short Papers)}, pages 1162--1172. Association for Computational
  Linguistics.

\bibitem[{Radford et~al.(2019)Radford, Wu, Child, Luan, Amodei, and
  Sutskever}]{radford2019language}
Alec Radford, Jeff Wu, Rewon Child, David Luan, Dario Amodei, and Ilya
  Sutskever. 2019.
\newblock Language models are unsupervised multitask learners.

\bibitem[{Riedel et~al.(2010)Riedel, Yao, and McCallum}]{riedel2010modeling}
Sebastian Riedel, Limin Yao, and Andrew McCallum. 2010.
\newblock \href {https://doi.org/10.1007/978-3-642-15939-8\_10} {Modeling
  relations and their mentions without labeled text}.
\newblock In \emph{Machine Learning and Knowledge Discovery in Databases,
  European Conference, {ECML} {PKDD} 2010, Barcelona, Spain, September 20-24,
  2010, Proceedings, Part {III}}, volume 6323 of \emph{Lecture Notes in
  Computer Science}, pages 148--163. Springer.

\bibitem[{Shang et~al.(2022)Shang, Huang, Sun, Wei, and
  Mao}]{Shang-pattern-relation-22}
Yuming Shang, Heyan Huang, Xin Sun, Wei Wei, and Xian{-}Ling Mao. 2022.
\newblock \href {https://doi.org/10.1016/j.ins.2021.10.047} {A pattern-aware
  self-attention network for distant supervised relation extraction}.
\newblock \emph{Inf. Sci.}, 584:269--279.

\bibitem[{Shannon(1948)}]{Shannon-entropy}
Claude~E. Shannon. 1948.
\newblock \href {https://doi.org/10.1002/j.1538-7305.1948.tb01338.x} {A
  mathematical theory of communication}.
\newblock \emph{Bell Syst. Tech. J.}, 27(3):379--423.

\bibitem[{Shelmanov et~al.(2021)Shelmanov, Tsymbalov, Puzyrev, Fedyanin,
  Panchenko, and Panov}]{shelmanov-etal-2021-certain}
Artem Shelmanov, Evgenii Tsymbalov, Dmitri Puzyrev, Kirill Fedyanin, Alexander
  Panchenko, and Maxim Panov. 2021.
\newblock \href {https://doi.org/10.18653/v1/2021.eacl-main.157} {How certain
  is your {T}ransformer?}
\newblock In \emph{Proceedings of the 16th Conference of the European Chapter
  of the Association for Computational Linguistics: Main Volume}, pages
  1833--1840, Online. Association for Computational Linguistics.

\bibitem[{Tan et~al.(2018)Tan, Wang, Xie, Chen, and Shi}]{tan2018deep}
Zhixing Tan, Mingxuan Wang, Jun Xie, Yidong Chen, and Xiaodong Shi. 2018.
\newblock \href {https://dl.acm.org/doi/abs/10.5555/3504035.3504639} {Deep
  semantic role labeling with self-attention}.
\newblock In \emph{Proceedings of the Thirty-Second AAAI Conference on
  Artificial Intelligence and Thirtieth Innovative Applications of Artificial
  Intelligence Conference and Eighth AAAI Symposium on Educational Advances in
  Artificial Intelligence}, AAAI'18/IAAI'18/EAAI'18. AAAI Press.

\bibitem[{Wang and Wang(2022)}]{wang-etal-2022iclr-ensemble}
Hongjun Wang and Yisen Wang. 2022.
\newblock \href {https://openreview.net/forum?id=oU3aTsmeRQV} {Self-ensemble
  adversarial training for improved robustness}.
\newblock In \emph{The Tenth International Conference on Learning
  Representations, {ICLR} 2022, Virtual Event, April 25-29, 2022}.
  OpenReview.net.

\bibitem[{Wang et~al.(2019)Wang, Liu, Wang, Luan, and Sun}]{wang-etal-2019}
Shuo Wang, Yang Liu, Chao Wang, Huanbo Luan, and Maosong Sun. 2019.
\newblock \href {https://doi.org/10.18653/v1/D19-1073} {Improving
  back-translation with uncertainty-based confidence estimation}.
\newblock In \emph{Proceedings of the 2019 Conference on Empirical Methods in
  Natural Language Processing and the 9th International Joint Conference on
  Natural Language Processing, {EMNLP-IJCNLP} 2019, Hong Kong, China, November
  3-7, 2019}, pages 791--802. Association for Computational Linguistics.

\bibitem[{Xiao and Wang(2019)}]{Xiao-aaai-19}
Yijun Xiao and William~Yang Wang. 2019.
\newblock \href {https://doi.org/10.1609/aaai.v33i01.33017322} {Quantifying
  uncertainties in natural language processing tasks}.
\newblock In \emph{The Thirty-Third {AAAI} Conference on Artificial
  Intelligence, {AAAI} 2019, The Thirty-First Innovative Applications of
  Artificial Intelligence Conference, {IAAI} 2019, The Ninth {AAAI} Symposium
  on Educational Advances in Artificial Intelligence, {EAAI} 2019, Honolulu,
  Hawaii, USA, January 27 - February 1, 2019}, pages 7322--7329. {AAAI} Press.

\bibitem[{Yu et~al.(2020)Yu, Zhang, Shu, Liu, Wang, Wang, and Li}]{yu2019joint}
Bowen Yu, Zhenyu Zhang, Xiaobo Shu, Tingwen Liu, Yubin Wang, Bin Wang, and
  Sujian Li. 2020.
\newblock \href {https://doi.org/10.3233/FAIA200356} {Joint extraction of
  entities and relations based on a novel decomposition strategy}.
\newblock In \emph{{ECAI} 2020 - 24th European Conference on Artificial
  Intelligence, 29 August-8 September 2020, Santiago de Compostela, Spain,
  August 29 - September 8, 2020 - Including 10th Conference on Prestigious
  Applications of Artificial Intelligence {(PAIS} 2020)}, volume 325 of
  \emph{Frontiers in Artificial Intelligence and Applications}, pages
  2282--2289. {IOS} Press.

\bibitem[{Zheng et~al.(2017)Zheng, Wang, Bao, Hao, Zhou, and
  Xu}]{zheng-etal-2017-joint}
Suncong Zheng, Feng Wang, Hongyun Bao, Yuexing Hao, Peng Zhou, and Bo~Xu. 2017.
\newblock \href {https://doi.org/10.18653/v1/P17-1113} {Joint extraction of
  entities and relations based on a novel tagging scheme}.
\newblock In \emph{Proceedings of the 55th Annual Meeting of the Association
  for Computational Linguistics (Volume 1: Long Papers)}, pages 1227--1236,
  Vancouver, Canada. Association for Computational Linguistics.

\bibitem[{Zhong and Chen(2021)}]{zhong-chen-2021-frustratingly}
Zexuan Zhong and Danqi Chen. 2021.
\newblock \href {https://doi.org/10.18653/v1/2021.naacl-main.5} {A
  frustratingly easy approach for entity and relation extraction}.
\newblock In \emph{Proceedings of the 2021 Conference of the North American
  Chapter of the Association for Computational Linguistics: Human Language
  Technologies}, pages 50--61, Online. Association for Computational
  Linguistics.

\bibitem[{Zhou et~al.(2020)Zhou, Yang, Wong, Wan, and
  Chao}]{zhou-etal-2020-uncertainty}
Yikai Zhou, Baosong Yang, Derek~F. Wong, Yu~Wan, and Lidia~S. Chao. 2020.
\newblock \href {https://doi.org/10.18653/v1/2020.acl-main.620}
  {Uncertainty-aware curriculum learning for neural machine translation}.
\newblock In \emph{Proceedings of the 58th Annual Meeting of the Association
  for Computational Linguistics}, pages 6934--6944, Online. Association for
  Computational Linguistics.

\end{thebibliography}
\bibliographystyle{acl_natbib}

\appendix

\end{document}